\documentclass[lettersize,journal]{IEEEtran}

\IEEEoverridecommandlockouts                              

\makeatletter
\let\NAT@parse\undefined
\makeatother

\usepackage[switch]{lineno}

\usepackage{amsmath}
\usepackage{amsfonts}
\usepackage{amssymb}
\usepackage{mathtools}
\usepackage{pifont}          

\usepackage{booktabs}
\usepackage{tabularx}
\usepackage{multirow}
\usepackage{threeparttable}
\usepackage{makecell}
\usepackage{adjustbox}
\usepackage{graphicx}
\usepackage{float}           
\usepackage{caption}
\usepackage{stfloats}
\usepackage{balance}         

\usepackage[table, dvipsnames]{xcolor}
\usepackage{xstring}
\usepackage{xparse}
\usepackage{etoolbox}

\usepackage{fancyhdr}
\usepackage{enumitem}

\usepackage{textcomp}
\usepackage{verbatim}
\usepackage{comment}

\usepackage{xargs}           
\usepackage[colorinlistoftodos,prependcaption,textsize=tiny]{todonotes}
\usepackage{listings}

\usepackage[noadjust]{cite}
\usepackage{url}
\usepackage{hyperref}
\usepackage{cleveref} 

\hypersetup{
    colorlinks=true,
    citecolor=Green,
    linkcolor=NavyBlue,
    filecolor=OrangeRed,
    urlcolor=OrangeRed,
}

\newcommand{\ModulePalette}{A}

\IfStrEqCase{\ModulePalette}{%
  {A}{\definecolor{ScenarioColor}{HTML}{063480}%
      \definecolor{ScenarioFeaturesColor}{HTML}{018077}%
      \definecolor{CriticalProbeColor}{HTML}{7D013F}%
      \definecolor{ScenarioScoresColor}{HTML}{7C4900}}%
  {B}{\definecolor{ScenarioColor}{HTML}{008673}%
      \definecolor{ScenarioFeaturesColor}{HTML}{7B6000}%
      \definecolor{CriticalProbeColor}{HTML}{780121}%
      \definecolor{ScenarioScoresColor}{HTML}{402B77}}%
  {C}{\definecolor{ScenarioColor}{HTML}{004998}%
      \definecolor{ScenarioFeaturesColor}{HTML}{04834A}%
      \definecolor{CriticalProbeColor}{HTML}{780121}%
      \definecolor{ScenarioScoresColor}{HTML}{64155F}}%
}[\errmessage{Unknown \ModulePalette: use A, B, or C}]

\colorlet{ScenarioFill}{ScenarioColor!20}
\colorlet{ScenarioFeaturesFill}{ScenarioFeaturesColor!20}
\colorlet{CriticalProbeFill}{CriticalProbeColor!20}
\colorlet{ScenarioScoresFill}{ScenarioScoresColor!20}

\definecolor{WomdColor}{HTML}{1B82E2}
\definecolor{NuScenesColor}{HTML}{47C9B9}
\definecolor{NuPlanColor}{HTML}{30B3A1}
\definecolor{ArgoverseColor}{HTML}{C98C16}

\lstdefinestyle{pythonschema}{
    language=Python,
    basicstyle=\ttfamily\scriptsize,
    keywordstyle=\color{RoyalBlue!70!black}\bfseries,
    stringstyle=\color{green!50!black},
    commentstyle=\color{gray}\itshape,
    showstringspaces=false,
    breaklines=true,
    captionpos=b,
    xleftmargin=0.5em,
    xrightmargin=0.5em,
    classoffset=1,
    morekeywords={
        ABC, Protocol, BaseModel, DictConfig,
        AgentType, AgentPairType, TrajectoryType, AgentTrajectory,
        BaseVisualizer, Enum, SupportedPanes,
        Float32NDArray1D, Float32NDArray2D, Float32NDArray3D, Float32NDArray4D,
        Int32NDArray1D, Int32NDArray2D, BooleanNDArray1D, NonNegativeInt,
    },
    keywordstyle=\color{Black}\bfseries,
    classoffset=2,
    morekeywords={
        Scenario, AgentData, ScenarioMetadata, TracksToPredict, StaticMapData, DynamicMapData,
    },
    keywordstyle=\color{ScenarioColor}\bfseries,
    classoffset=3,
    morekeywords={BaseFeature, ScenarioFeatures, Individual, Interaction, InteractionStatus},
    keywordstyle=\color{ScenarioFeaturesColor}\bfseries,
    classoffset=4,
    morekeywords={ScenarioScores, Score, BaseScorer},
    keywordstyle=\color{ScenarioScoresColor}\bfseries,
    classoffset=5,
    morekeywords={ProbeType, ProbeValidity, ProbeFn, CriticalityResult, CriticalProbe, ScenarioProbe, CriticalityMetric, ProbeValidator, BehaviorProber},
    keywordstyle=\color{CriticalProbeColor}\bfseries,
}

\newcommand{\idest}{i.e., }
\newcommand{\exempli}{e.g., }

\newcommand{\paragraphbf}[1]{\vspace{0.2cm}\noindent\textbf{#1.}}

\newcommand{\framework}{{\textcolor{Black!70}{\textsc{ScenarioCharacterization}}}}
\newcommand{\scenario}{{\textcolor{ScenarioColor}{\texttt{Scenario}}}}
\newcommand{\features}{{\textcolor{ScenarioFeaturesColor}{\texttt{ScenarioFeatures}}}}
\newcommand{\probe}{{\textcolor{CriticalProbeColor}{\texttt{ScenarioProbe}}}}
\newcommand{\scores}{{\textcolor{ScenarioScoresColor}{\texttt{ScenarioScores}}}}

\newcommand{\vehicle}{\text{vehicle}}
\newcommand{\vehicleC}{{\textcolor{Black!80}{\text{vehicle}}}}
\newcommand{\pedestrian}{\text{pedestrian}}
\newcommand{\pedestrianC}{{\textcolor{Magenta!85}{\text{pedestrian}}}}
\newcommand{\cyclist}{\text{cyclist}}
\newcommand{\cyclistC}{{\textcolor{ForestGreen!85}{\text{cyclist}}}}
\newcommand{\egoagent}{\text{ego-agent}}
\newcommand{\egoagentC}{{\textcolor{RoyalBlue!85}{\text{ego-agent}}}}
\newcommand{\relevant}{\text{relevant}}
\newcommand{\relevantC}{{\textcolor{Orange!85}{\text{relevant}}}}

\newcommand{\womd}{\textsc{WOMD}}
\newcommand{\womdC}{{\textcolor{WomdColor}{\textsc{WOMD}}}}
\newcommand{\nuscenes}{\textsc{nuScenes}}

\newcommand{\nuplan}{\textsc{nuPlan}}
\newcommand{\nuplanC}{{\textcolor{NuPlanColor}{\textsc{nuPlan}}}}
\newcommand{\argoverse}{\textsc{Argoverse2}}
\newcommand{\argoverseC}{{\textcolor{ArgoverseColor}{\textsc{Argoverse2}}}}

\newcommand{\safeshift}{{\textcolor{Black!95}{\textsc{SafeShift}}}}
\newcommand{\unitraj}{{\textcolor{Black!95}{\textsc{UniTraj}}}}
\newcommand{\scenarionet}{{\textcolor{Black!95}{\textsc{ScenarioNet}}}}
\newcommand{\trajdata}{{\textcolor{Black!95}{\textsc{trajdata}}}}
\newcommand{\commonroad}{{\textcolor{Black!95}{\textsc{CommonRoad}}}}
\newcommand{\scenic}{{\textcolor{Black!95}{\textsc{Scenic}}}}
\newcommand{\metadrive}{{\textcolor{Black!95}{\textsc{MetaDrive}}}}
\newcommand{\causalagents}{{\textcolor{Black!95}{\textsc{CausalAgents}}}}

\newcommandx{\unsure}[2][1=]{\todo[linecolor=red,backgroundcolor=red!25,bordercolor=red,#1]{#2}}
\newcommandx{\change}[2][1=]{\todo[linecolor=blue,backgroundcolor=blue!25,bordercolor=blue,#1]{#2}}
\newcommandx{\info}[2][1=]{\todo[linecolor=OliveGreen,backgroundcolor=OliveGreen!25,bordercolor=OliveGreen,#1]{#2}}
\newcommandx{\improvement}[2][1=]{\todo[linecolor=Plum,backgroundcolor=Plum!25,bordercolor=Plum,#1]{#2}}
\newcommandx{\thiswillnotshow}[2][1=]{\todo[disable,#1]{#2}}
\def\HiLiYellow{\leavevmode\rlap{\hbox to \hsize{\color{yellow!20}\leaders\hrule height .8\baselineskip depth .4ex\hfill}}}

\newif\ifshowstatshorizontal
\showstatshorizontaltrue

\fancypagestyle{firststyle}{
  \fancyhf{}

}

\fancypagestyle{secondstyle}{
  \fancyhf{}
  \fancyhead[R]{\footnotesize \thepage}

}

\graphicspath{{figures/}}
\newcommand{\includegraphicsdpi}[3]{
    \pdfimageresolution=#1  
    \includegraphics[#2]{#3}
    \pdfimageresolution=72  
}

\title{\framework: A Modular Toolkit for Characterizing Safety across Trajectory Datasets}

\author{
Ingrid Navarro,$^{1,2}$ 
Yutong Duan,$^{3}$ 
Jonathan Francis$^{1,4,\dag}$ 
and Jean Oh$^{1,\dag}$
\thanks{$^{1}$Robotics Institute, School of Computer Science, Carnegie Mellon University. {\tt\scriptsize \{ingridn, jeanoh\}@cs.cmu.edu}}
\thanks{$^{2}$Work done as part of an internship at Stack AV.}
\thanks{$^{3}$Stack AV}
\thanks{$^{4}$Bosch Center for Artificial Intelligence: {\tt\scriptsize jon.francis@us.bosch.com}}
\thanks{$^{\dag}$Equal Advising.}
}

\let\oldtwocolumn\twocolumn
\renewcommand\twocolumn[1][]{%
    \oldtwocolumn[{#1}{
    \begin{center}
           \captionsetup{hypcap=false}
           \vspace{-0.65cm}
           \includegraphicsdpi{200}{width=0.98\textwidth}{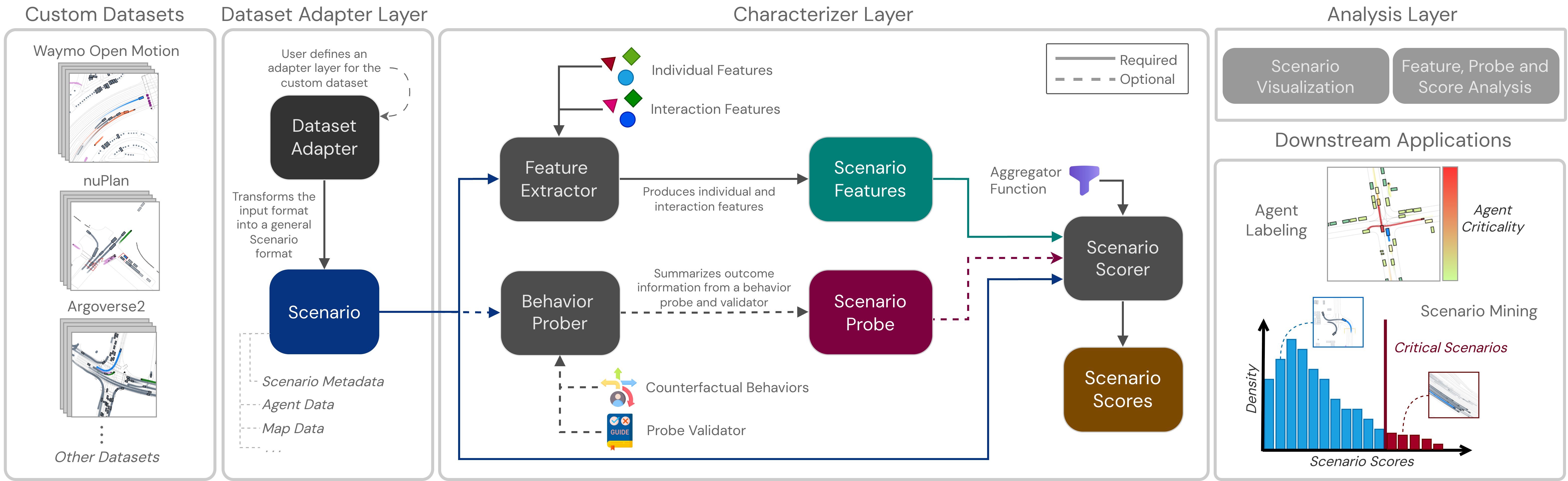}
           \captionof{figure}{Overview of~\framework, a framework for automated, dataset-agnostic profiling of driving scenarios in trajectory datasets. Colored blocks denote the schemas that carry data between layers, and match the color used for \scenario, \features, \probe, and \scores~throughout the text. Solid arrows mark required paths, dashed arrows optional ones.}
           \label{fig:overview}
        \end{center}
    }]
}

\begin{document}
\maketitle

\begin{abstract}
We introduce \framework, an open-source framework for automated, dataset-agnostic profiling of driving scenarios in trajectory datasets. Our framework is packaged as a modular, configuration-driven pipeline of three layers: a \textit{dataset adapter} that maps custom datasets onto an open \scenario~representation, a \textit{characterizer} that performs feature extraction, behavior probing, and criticality scoring at scenario and agent levels, and an \textit{analysis} layer for scenario visualization and feature, score, and probe analyses. Because the layers communicate only through Pydantic-validated schemas composed via configurations, a new dataset can easily plug in without rewriting the characterization and analysis stack.

We describe the design and APIs, show example outputs on Waymo Open Motion, Argoverse2, and nuPlan, and discuss downstream uses of the approach. The framework is available at \href{https://github.com/navarrs/ScenarioCharacterization}{github.com/navarrs/ScenarioCharacterization}.
\end{abstract}

\section{Introduction} 
\label{sec:introduction}

Trajectory datasets have become a backbone of modern autonomous-driving (AD) development. Benchmarks such as the Waymo Open Motion Dataset (\womd)~\cite{ettinger2021large}, \argoverse~\cite{wilson2021argoverse}, and \nuplan~\cite{karnchanachari2024nuplan} provide large collections of recorded urban interactions that power much of the open research in trajectory prediction, planning, and simulation. Yet recent work~\cite{stoler2024safeshift, bahari2022vehicle} shows that models trained on these datasets remain brittle under unseen conditions, and highlights the value of systematically profiling scenarios to surface meaningful, rare interactions.

Such analyses, however, are typically constrained to dataset-specific settings. \safeshift~\cite{stoler2024safeshift}, for example, is built around the \womd\ scenario format. Porting it to \argoverse\ or \nuplan\ requires adapting its logic to each dataset's distinct representation. This makes insights hard to transfer and slows systematic, cross-dataset study.

\framework~addresses this gap with a generalizable, automated framework that lifts scenario characterization out of a single setting into a reusable library. Its central design choice is to normalize every dataset into our \scenario~representation, defined by documented Pydantic~\cite{pydantic} data validation schemas, and to have each downstream stage communicate only through schemas. Because characterization and analysis stages operate on these representations rather than on any source format, integrating a new dataset largely reduces to writing a lightweight adapter that populates the \scenario~representation. Downstream stages are designed to run unchanged. These stages are in turn assembled via configuration groups~\cite{Yadan2019Hydra}.

By decoupling characterization from any single dataset's format, our framework is intended to support a range of cross-dataset downstream applications. We envision it enabling \textit{scenario mining and curation}, where criticality scores surface the long tail of rare or safety-critical situations for targeted training and evaluation~\cite{stoler2024safeshift, ding2025surprise, yang2024hard}; \textit{simulation and scenario generation}, where characterization guides synthesis of scenarios matching or exceeding the difficulty of recorded data~\cite{ding2023survey, huang2025cadre}; \textit{model stress-testing}, where stratifying performance by scenario difficulty exposes failure modes that aggregate metrics hide~\cite{stoler2024safeshift}; \textit{cross-dataset benchmarking}, since the same pipeline runs on any compliant dataset~\cite{feng2024unitraj}; and \textit{labeling and auxiliary supervision}, where raw and categorized scores serve as supervisory or curriculum-learning signals~\cite{stoler2024safeshift}.

To summarize, this report contributes the following:
\begin{itemize}
    \item We present a configuration-driven scenario profiling pipeline that integrates characterization components from prior work~\cite{stoler2024safeshift, feng2024unitraj, glasmacher2022automated} into a single reusable library. 
    \item We introduce a schema-based model that maps heterogeneous datasets onto a unified representation, so downstream characterization stages operate independently of the source benchmark. 
    \item We show worked examples on \womd, \argoverse, and \nuplan\ and demonstrate the framework across two downstream uses: \textit{scenario mining} and \textit{agent labeling}.
\end{itemize}

\section{Related Work}
\label{sec:related_work}

\subsection{Trajectory Dataset Analysis} 
\label{ssec:trajectory_dataset_analysis}

Many recent works focus on surfacing critical or interactive behaviors within existing datasets. For instance, Glasmacher et al.~\cite{glasmacher2022automated} derive a taxonomy of detection types from scenario tracks and combine them hierarchically into interaction, anomaly, and relevance scores for dataset comparison, demonstrated on naturalistic drone recordings~\cite{krajewski2018highd}. \safeshift~\cite{stoler2024safeshift} likewise introduces a hierarchical score, fusing individual and social features into a safety measure to create distribution-aware splits. Ding et al.~\cite{ding2025surprise} propose \textit{surprise potential}, a counterfactual metric that quantifies interactivity by measuring an AV's potential to elicit surprising behavior from others. The \causalagents~\cite{sun2024causalagents} benchmark annotates agents in a \womd~subset as causal or non-causal to the AV's behavior, using these labels to perturb scenarios and probe motion-forecasting robustness. 

We build on these ideas but differ in exposing feature axes, aggregators, probes, and downstream analyses as configurable extension points, decoupling each methodology from the benchmark it was originally introduced on. A parallel line of work, which we do not address here, focuses on \textit{generating} critical or interactive scenarios rather than characterizing recorded ones~\cite{ding2023survey, huang2025cadre, stoler2025seal}.

\subsection{Unified Frameworks for Trajectory Data} 
\label{ssec:unified_frameworks_trajectory_data}

Various frameworks lift trajectory datasets out of single-benchmark settings into reusable libraries with a common \textit{representation}. \scenarionet~\cite{li2023scenarionet} defines a unified scenario description format that ingests heterogeneous driving datasets and pairs it with the \metadrive~\cite{li2023metadrive} simulator for tasks such as scenario replay, imitation and reinforcement learning, and AD-stack testing. \trajdata~\cite{ivanovic2023trajdata} provides a unified data-loader interface across pedestrian and AV trajectory datasets and uses it for a cross-dataset comparison. \unitraj~\cite{feng2024unitraj} likewise unifies multiple datasets and prediction models behind a common pipeline to study generalization and data-scaling effects on trajectory prediction. A related line instead standardizes benchmark \textit{specification}. \commonroad~\cite{althoff2017commonroad} couples a scenario to a vehicle model and cost function under a unique reproducibility ID, while \scenic~\cite{fremont2019scenic} provides a probabilistic language for describing scenario distributions from which concrete scenes can be sampled.

We complement these methods by focusing on the axis of \textit{scenario characterization}. Our unified \scenario~representation is similar to the formats of \scenarionet~and \trajdata, and, like \commonroad, we support reproducible specification. What differs in our work is that our machinery is oriented toward profiling, mining, and labeling scenarios that can feed into simulation, prediction, and other tasks.

\section{Framework Overview}
\label{sec:framework}

\framework, shown in \Cref{fig:overview}, is organized into three layers. 
The \textbf{Dataset Adapter} (\Cref{sec:dataset_adapter}) translates a custom dataset into our unified \scenario~representation. 
The \textbf{Characterizer} layer (\Cref{sec:characterizer}) consumes a \scenario~and produces structured descriptors through three components: a \textit{Feature Extractor} which computes features across various axes, a \textit{Scorer} which aggregates those features into per-agent and per-scenario scores, and an optional \textit{Behavior Prober} which probes latent criticality via counterfactual behavior. 
The \textbf{Analysis} layer (\Cref{sec:analysis}) consumes the outputs of these layers to support scenario visualization and distribution analysis over features, probes, and scores.

\Cref{tab:api} summarizes the extension point each component exposes for implementation. Each is extended by subclassing a base class, implementing the corresponding entry point, and registering the expected output format. The following subsections walk through each layer with schema listings\footnote{For readability, the listings are simplified. We omit optional fields, exact array types, and occasionally use a clearer name than the corresponding symbol in the codebase, to which we refer the reader for full details.}.

\begin{table}[h]
\caption{Framework API Overview. Each component is extended by subclassing its base class and implementing the corresponding entry point. 
}
\label{tab:api}
\centering
\resizebox{\columnwidth}{!}{%
\begin{tabular}{@{}lll@{}}
\toprule
\textbf{Component Base Class} & \textbf{Entry point} & \textbf{Output} \\
\midrule
\texttt{BaseDataset} (\ref{sec:dataset_adapter}) & \texttt{transform(Raw Scenario)} & \scenario \\
\texttt{BaseFeature} (\ref{ssec:feature}) & \texttt{compute(\scenario)} & \features \\
\texttt{BaseScorer} (\ref{ssec:scorer}) & \texttt{compute(\scenario, \features)} & \scores \\
\texttt{BehaviorProber} (\ref{ssec:prober}) & \texttt{probe(\scenario)} & \probe \\
\texttt{BaseVisualizer} (\ref{ssec:scenario_viz}) & \texttt{visualize(\scenario)} & Rendered Scenario \\
\bottomrule
\end{tabular}
}
\end{table}

\section{Dataset Adapter Layer}
\label{sec:dataset_adapter}

The adapter layer is the only layer that a user needs to implement for their dataset. Its purpose is to map a dataset's native format onto a \scenario~(\Cref{lst:scenario}), the representation on which the rest of the framework operates. Like all others used in this framework, it builds upon Pydantic's \texttt{BaseModel}, which defines a schema and validates incoming data at runtime, ensuring that malformed inputs are caught before the characterization stage.

The \scenario~schema bundles complementary schemas capturing different components of a driving episode: \texttt{ScenarioMetadata} carries identifiers, timing information, and the threshold values used by downstream features. \texttt{AgentData} describes the trajectories, types, and relevance of every agent in the scene. \texttt{TracksToPredict} optionally flags the subset of agents targeted by prediction, navigation, or generation. \texttt{StaticMapData} includes static map information such as lanes, road geometry, regulatory elements, and derived conflict points. \texttt{DynamicMapData} includes dynamic points, such as traffic lights.

\begin{lstlisting}[style=pythonschema, caption={\texttt{\scenario} schema. We refer the reader to the codebase for further details on each of the sub-schemas.}, label={lst:scenario}]
from pydantic import BaseModel

class Scenario(BaseModel):
    metadata: ScenarioMetadata
    agent_data: AgentData
    tracks_to_predict: TracksToPredict 
    static_map_data: StaticMapData
    dynamic_map_data: DynamicMapData 
\end{lstlisting}



\paragraphbf{Custom Datasets} Integrating a new dataset reduces to writing a \texttt{BaseDataset} subclass that consumes the dataset's native format and produces \scenario~instances. Every downstream component then works without modification.

\section{Characterizer Layer}
\label{sec:characterizer} 

The characterizer layer builds on the hierarchical scenario-scoring algorithms of~\cite{stoler2024safeshift, glasmacher2022automated}, generalizing them so that each stage is selected and configured independently of the dataset it runs on. Each component is driven by a structured configuration specifying which features to extract (\Cref{ssec:feature}), which scoring function to use (\Cref{ssec:scorer}), and which behavior probe to apply (\Cref{ssec:prober}). New specifications can be developed, integrated, and selected through these configurations with minimal boilerplate. 

\subsection{Feature Extractor}
\label{ssec:feature}

The feature extractor consumes a \scenario~and produces descriptors that characterize an agent's behavior in its environment and the potential difficulty it may induce for surrounding agents. We organize these descriptors into \textit{feature axes}, \idest families of related quantities that share a common scope, such as the behavior of a single agent in isolation, or the interactive behavior between pairs of agents. The descriptors produced by active axes are gathered into the \features~schema (\Cref{lst:scenario_features}).

\vspace{0.1cm}\subsubsection{Feature Extraction API}
Each feature axis subclasses \texttt{BaseFeature} (\Cref{tab:api}) and is constructed from a structured configuration, keeping the framework agnostic to the number, scope, or semantics of the axes a given study chooses to use.

\vspace{0.1cm}\subsubsection{Supported Feature Extractors}
We support two families of feature extractors summarized in \Cref{tab:features}:

\paragraphbf{Individual Features} These descriptors, adopted from \cite{stoler2024safeshift, glasmacher2022automated, makansi2021exposing, feng2024unitraj}, characterize each agent individually along three dimensions: kinematic profile (speed, acceleration, jerk), environmental compliance (speed-limit deviation, waiting periods at conflict points), and trajectory complexity (trajectory type and Kalman-based prediction difficulty).

\begin{table}[!htbp]
\caption{List of low-level per-axis features.}
\label{tab:features}
\resizebox{\columnwidth}{!}{%
\begin{tabular}{ll}
\toprule
\textbf{Individual Features }    & \textbf{Interaction Features  }         \\
\midrule
Speed, Acceleration, Jerk~\cite{stoler2024safeshift}                                 & Time-to-Collision (TTC)~\cite{stoler2024safeshift, glasmacher2022automated}                 \\
Speed Limit Diff~\cite{stoler2024safeshift}                                          & Time Headway (THW)~\cite{stoler2024safeshift, glasmacher2022automated}                      \\
Waiting Period at Conflict Point~\cite{stoler2024safeshift, glasmacher2022automated} & Deceleration Rate to Avoid Crash (DRAC)~\cite{stoler2024safeshift, glasmacher2022automated} \\
Trajectory Type~\cite{feng2024unitraj}                                               & Minimum Time to Conflict Point (mTTCP)~\cite{stoler2024safeshift, glasmacher2022automated}  \\
Kalman Difficulty~\cite{feng2024unitraj, makansi2021exposing}                        & Collisions~\cite{stoler2024safeshift, glasmacher2022automated}                              \\
\bottomrule
\end{tabular}
}
\end{table}

\paragraphbf{Interaction\footnote{\safeshift~\cite{stoler2024safeshift} utilizes the word \textit{social} to refer to this axis.} Features} These descriptors, adopted from \cite{stoler2024safeshift, glasmacher2022automated, feng2024unitraj}, characterize how severely two agents could come into conflict. They include collision indicators alongside established surrogate safety measures~\cite{westhofen2023criticality}: minimum Time to Conflict Point (mTTCP), Time Headway (THW), Time to Collision (TTC), and Deceleration Rate to Avoid a Crash (DRAC). We annotate interacting pairs with a validity status and the types of agents involved, so that subsequent analyses can condition on the kind of interaction observed.

These axes are aggregated into a top-level container, \features, shown below. 
\begin{lstlisting}[style=pythonschema, caption={\features~container schema.}, label={lst:scenario_features}]
class Individual(BaseModel):
    valid_agents: Int32NDArray1D
    agent_types: list[AgentType]
    agent_trajectory_types: list[TrajectoryType]
    speed_limit_diff: Float32NDArray1D
    acceleration_profile: Float32NDArray1D
    waiting_period: Float32NDArray1D
    kalman_difficulty: Float32NDArray1D
    # Other features...
    
class Interaction(BaseModel):
    interaction_status: list[InteractionStatus]
    interaction_agent_types: list[AgentPairType]
    collision: Float32NDArray1D
    mttcp: Float32NDArray1D
    thw: Float32NDArray1D
    ttc: Float32NDArray1D
    drac: Float32NDArray1D
    # Other features...
    
class ScenarioFeatures(BaseModel):
    individual_features: Individual 
    interaction_features: Interaction 
    # Other axes...
\end{lstlisting}

\paragraphbf{Custom Feature Axes} Adding axes requires defining a schema with the descriptors it is responsible for and designating a slot in \features. The user decides its scope, descriptors, and any precomputed requirements.

\subsection{Scorer}
\label{ssec:scorer}

The scorer turns the descriptors produced by the \textit{Feature Extractor} and, optionally, the \textit{Behavior Prober} into agent-level and scenario-level scores. Building upon~\cite{stoler2024safeshift}, we organize this step around \textit{aggregator functions}: components that summarize one or more feature axes into a single scalar per agent and per scene. The outputs of the active aggregators are collected in the \scores~schema (\Cref{lst:scenario_scores}).

\vspace{0.1cm}\subsubsection{Score Extraction API}
Each aggregator subclasses \texttt{BaseScorer} (\Cref{tab:api}), which provides default utilities such as importance weighting, score binning, and expert-defined thresholds, so the aggregator functions focus on implementing scoring rules. As with the features API, the set of aggregators can be specified via declarative configurations.

\vspace{0.1cm}\subsubsection{Supported Score Aggregators}

The currently supported aggregators follow \cite{stoler2024safeshift}:

\paragraphbf{Individual Aggregator} For each agent in the scene, it produces an \textit{individual} score, computed from the agent's per-feature critical values clipped against feature-specific thresholds and combined via a per-feature importance-weighted sum. 

\paragraphbf{Interaction Aggregator} For each agent pair, it clips the pair's interaction features against feature-specific thresholds and combines them through a per-feature importance-weighted sum, yielding an \textit{interaction} score. The importance weights can be shaped to focus, \exempli on pairs including the \egoagent, or, if available, on relevant agents. 

\begin{lstlisting}[style=pythonschema, caption={\scores~container schema.}, label={lst:scenario_scores}]
class Score(BaseModel):
    agent_scores: Float32NDArray1D
    agent_scores_valid: BooleanNDArray1D
    scene_score: float
    
class ScenarioScores(BaseModel):
    individual_scores: Score
    interaction_scores: Score
    safeshift_scores: Score
\end{lstlisting}

\paragraphbf{SafeShift Aggregator} A hierarchical aggregator that combines the \textit{individual} and \textit{interaction} scores into a final \textit{safeshift} score weighted by proximity to a relevant agent (\exempli the \egoagent). We extend the \safeshift~aggregation with an additional multiplier that prioritizes interactions with vulnerable road users (VRUs). The corresponding scene-level value along each axis is the sum of its per-agent values normalized by the number of scored agents in the scene. We refer the reader to \cite{stoler2024safeshift, glasmacher2022automated} for the full formulation and for the origin of the per-feature thresholds and weights, which we re-calibrate from the observed feature distributions of the target dataset.

\paragraphbf{Custom Aggregators} A new aggregator decides which feature axes to draw from and how their values are weighted.


\subsection{Behavior Prober}
\label{ssec:prober}

This component probes a \scenario~by replacing one agent's future trajectory at a time with a counterfactual behavior. Then, it checks if the resulting scenario exhibits more criticality than the original. The goal is to surface latent criticality, \idest situations that are nominal as recorded but could have been critical under plausible behavioral changes. 

It is organized around three components: a \textit{counterfactual probe} that defines the alternative behavior, a \textit{probe validator} that scores the perturbed scenario and decides whether the perturbation is relevant, and a \textit{prober} that sweeps probes across agents and selects the \textbf{most impactful} one. Its output follows the \probe~schema (\Cref{lst:critical_probe}).

\vspace{0.1cm}\subsubsection{Behavior Probing API}
The prober exposes the entry point listed in \Cref{tab:api}. Both the probe (\exempli a constant velocity model) and the validator (\exempli a criticality measure) are configurable, allowing new perturbation models and validation criteria to be incorporated without altering the prober itself.

A scenario is probed by sweeping the active probe over (i) the ego agent vs.\ all others and (ii) each non-ego vs.\ the ego\footnote{Non-ego vs.\ non-ego pairs are excluded to reduce computational blowup, since each candidate requires perturbing and re-scoring the scenario.}. 
For each candidate, the prober perturbs the target agent's trajectory, recomputes the affected interaction features/scores and checks its validity. If no candidate is deemed valid, the algorithm returns an empty \probe~object, otherwise it retains the most impactful probe across the sweep, according to the configured conditions.

\begin{lstlisting}[style=pythonschema, caption={\texttt{CriticalProbe} schema.}, label={lst:critical_probe}]
class CriticalProbe(BaseModel):
    probed_agent_id: int
    probed_agent_trajectory: Float32NDArray2D
    is_ego_agent: bool
    probe_type: ProbeType
    criticality_time_idx: int

    # Scene scores before and after probing
    scene_score_before: float 
    scene_score_after: float 

    # Affected agents' scores before/after probing
    affected_agent_ids: list[int]
    affected_pair_scores_before: dict[str, float]
    affected_pair_scores_after: dict[str, float]

    # Additional critical probe metadata...
\end{lstlisting}







\subsubsection{Supported Probe and Validator}

We currently provide support for the following probe and validation functions:

\paragraphbf{Constant-velocity Probe} This probe replaces an agent's future frames with constant-velocity extrapolation starting one step after the time index defined in \texttt{current\_time\_index}, which is the last observed agent state. The motivation behind this baseline is to test whether a scenario remains safe under a possible deviation: an agent failing to react.

\paragraphbf{Score-based Validator} The role of the validator is to assess whether a probe is useful under a desired measure of criticality, \exempli a safety score or rule violation. Our validator scores the original and perturbed scenarios using the interaction-feature aggregator (\Cref{ssec:scorer}) and reports a per-pair score delta. We restrict the validator to this axis because rewriting one agent's trajectory changes the pairwise relationships that agent participates in while leaving the individual profiles of every other agent untouched, so the interaction score isolates the effect of the counterfactual. A probe is retained only if its delta exceeds a configurable threshold. Then, across all retained agent probes, the prober keeps the one with the largest delta, \idest the perturbation that increases criticality the most. 

\paragraphbf{Custom Probes and Validators} A new probe is a function satisfying the probe signature; a new validator is a callable that scores before/after pairs and returns a delta. The sweeping strategy can be reused as-is, so a user focuses only on the perturbation model or the criticality metric.

\section{Analysis Layer}
\label{sec:analysis}

The following subsections describe the visualization (\Cref{ssec:scenario_viz}) and characterization analysis (\Cref{ssec:characterization_analysis}) capabilities of this layer. We showcase example outputs on \womd, \argoverse, and \nuplan\footnote{Our released code also provides a \nuscenes~\cite{caesar2020nuscenes} adapter, but we exclude it from comparison because it contains fewer annotated scenes than the sample size used for our analyses.}.

\begin{figure*}[t]
    \centering
    \includegraphics[width=0.96\textwidth]{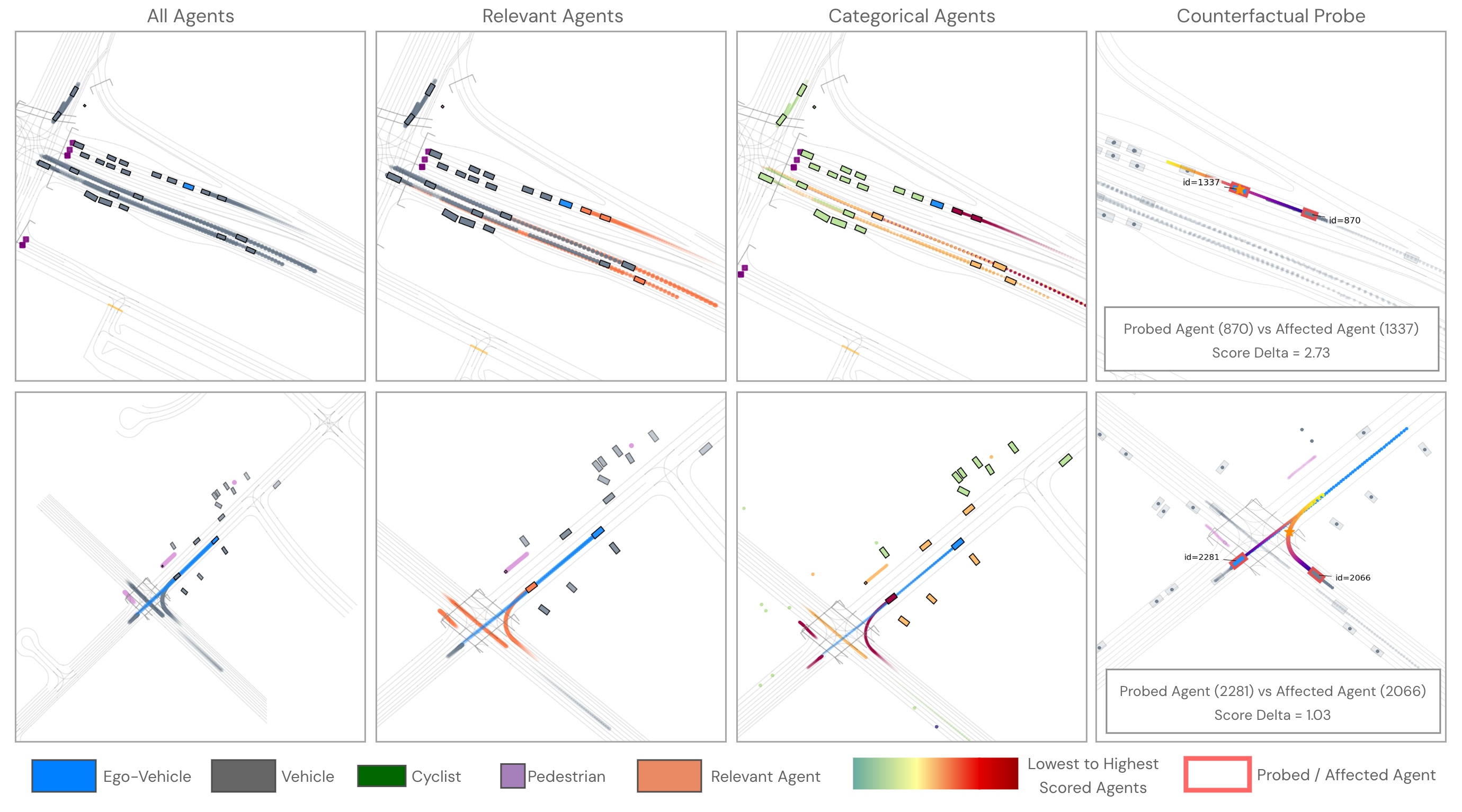}
    \caption{Supported visualizations, shown on two \womd~\cite{ettinger2021large} scenarios. \textbf{Top}: the ego waits at a traffic light while \relevant~agents approach and stop behind it. \textbf{Bottom}: the ego approaches an intersection as a \cyclist~initiates a left turn across its path. 
    }
    \label{fig:scenario_viz}
\end{figure*}

\subsection{Experimental Setup}
\label{ssec:experimental_setup}

For each dataset in our experiments, we characterize 5,000 temporally disjoint scenarios drawn uniformly under a fixed seed from a per-dataset pool. The agent statistics of these samples are summarized in \Cref{tab:experimental_setup}. 
To make the datasets comparable, we standardize them along two axes:

\begin{table}[!htpb]
    \centering
    \small
    \caption{Individual trajectories and interactions per dataset sample, reported as the per-scenario interquartile mean with the sample total in parentheses. 
    Interaction columns only count \textit{ego pairs}; global interaction counts are available in the code.
    }
    \resizebox{\columnwidth}{!}{%
    \begin{tabular}{lcccccc}
        \toprule
        & \multicolumn{3}{c}{\textbf{Trajectories}} & \multicolumn{3}{c}{\textbf{Ego interactions}} \\
        \cmidrule(lr){2-4} \cmidrule(lr){5-7}
        \textbf{Dataset} & Vehicles & Pedestrians & Cyclists & Ego-Vehicle & Ego-Pedestrian & Ego-Cyclist \\
        \midrule
        \womd~\cite{ettinger2021large}          & 49.2 (290k) & 3.4 (37k) & 0.0 (2.6k) & 31.5 (215k) & 0.8 (18k) & 0.0 (2.1k) \\
        \argoverse~\cite{wilson2021argoverse}   & 37.8 (202k) & 2.8 (25k) & 0.5 (6.0k) & 25.8 (144k) & 1.1 (17k) & 0.2 (4.5k) \\
        \nuplan~\cite{karnchanachari2024nuplan} & 32.6 (174k) & 21.0 (194k) & 0.3 (3.2k) & 20.6 (127k) & 3.5 (112k) & 0.2 (2.5k) \\
        \bottomrule
    \end{tabular}
    }
    \label{tab:experimental_setup}
\end{table}

\begin{itemize}
    \item \textit{Sampling Rate}: All scenarios are brought to a common 10\,Hz sampling rate.
    We maintain the scenario horizons of each dataset, \idest 91 timesteps (9.1\,s) for \womd, 110 (11.0\,s) for \argoverse, and 60 (6.0\,s) for \nuplan. 
    \item \textit{Agent Taxonomy}: We map the categories of \argoverse~and \nuplan~onto the \womd~types (\vehicle, \pedestrian, \cyclist), folding each dataset's vehicle-like and bicycle-like classes into \vehicle~and \cyclist, retaining \pedestrian~as-is, and dropping the remaining non-traffic classes into a residual \textit{other} type excluded from analysis. 
\end{itemize}
The per-dataset scenario selection and standardization procedure are documented in our codebase. Neither standardization is a framework constraint, \idest horizon, sampling rate, and agent taxonomy are per-dataset configuration. We apply the standardization only to align the cross-dataset comparison. 
 
Note that both characterizing all agent pairs and characterizing only those that include the \egoagent~are supported. For simplicity, the analysis that follows, \idest interaction extraction, aggregation, and probing, uses the latter. 
Finally, in the figures that follow, agent types are color-coded as: \vehicleC, \pedestrianC, \cyclistC, the \egoagentC, and \relevantC\footnote{Dataset users may have different criteria for labeling an agent as relevant, \exempli if they are critical to the ego or challenging for prediction. Our analysis uses \womd's \texttt{TracksToPredict}. }~agents. Datasets follow: \womdC, \argoverseC, and \nuplanC.

\subsection{Scenario Visualization}
\label{ssec:scenario_viz}

The visualization utility renders a \scenario~and, optionally, its characterizer outputs for inspection. It is built around two configurable axes of variation: \textit{visualizer types} and \textit{panes} that control how time and objects are rendered, respectively.

\subsubsection{Scenario Visualization API} 

We provide a configurable base class, \texttt{BaseVisualizer} (\Cref{tab:api}), which provides reusable plotting utilities to visualize map data and agent sequences and rendering options such as colors, time scale, and ego-level zoom, etc.

\paragraphbf{Visualizer Types} Our package supports two visualizer types, \textit{static} and \textit{animated}, that share the same plotting primitives but differ in how they render time. 

\vspace{0.1cm}\subsubsection{Supported Visualization Panes} Panes define how objects in the scenario will be visualized. Multiple panes can be composed into a single figure. We currently support: 
\begin{itemize}
    \item \textit{All Agents}: draws every agent in the scene following the color scheme in \Cref{ssec:experimental_setup}.
    \item \textit{Relevant Agents}: follows the above scheme and additionally highlights agents flagged as \relevant~(if available).
    \item \textit{Scored/Categorical Agents}: colors agents in a green to red spectrum based on criticality scores.
    \item \textit{Counterfactual Probes}: highlights probed and affected agents.
\end{itemize}
\noindent\Cref{fig:scenario_viz} illustrates the supported panes on two interactions.

\paragraphbf{Custom Visualizers and Panes} A new pane requires a rendering routine along with its configurable specifications.

\begin{figure*}[!t]
    \centering
    \includegraphics[width=0.96\textwidth]{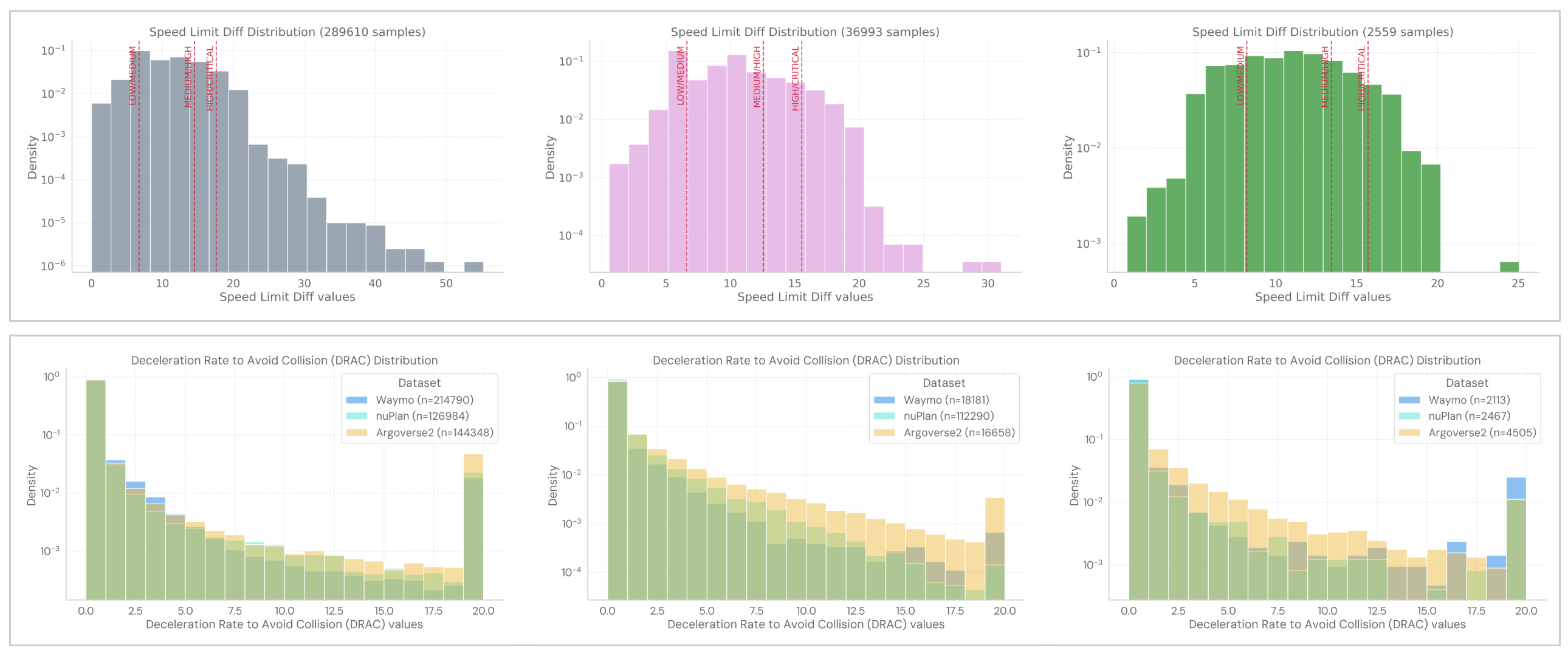}
    \caption{Feature distribution analysis examples. \textbf{Top}: within-dataset comparison on \womd, showing the \textit{speed-limit difference} distributions for vehicles (left), pedestrians (center) and cyclists (right). The red dashed lines show the 25th, 75th and 90th percentile values. \textbf{Bottom}: cross-dataset comparison across \womd, \nuplan, and \argoverse, showing \textit{DRAC} distributions for ego-vehicle (left), ego-pedestrian (center) and ego-cyclist (right) pairs. Both rows use a logarithmic $y$-axis.}
    \label{fig:feature_analysis}
\end{figure*}

\subsection{Characterizer Analysis}
\label{ssec:characterization_analysis}

We provide analysis tools that turn characterization outputs into dataset-level, per-agent-type distributional comparisons.

\vspace{0.1cm}\subsubsection{Feature analysis example}
\Cref{fig:feature_analysis} presents examples from our feature-level analysis. The top row compares the distribution of the \textit{speed limit difference} across agent types on \womd, with percentile boundaries shown as vertical dashed lines.
Vehicles (left) exhibit a right-skewed distribution that peaks near 10 and decays into a long tail beyond 50, whereas pedestrians (center) and cyclists (right) follow a flatter distribution concentrated between 5 and 20.
Because pedestrian and cyclist speeds are small and nearly constant relative to posted limits, their speed limit difference is dominated by the road limit itself; for vehicles, by contrast, the difference is more informative about behavior.

The bottom row extends the comparison across datasets for the DRAC feature across ego-vehicle (left), ego-pedestrian (center), and ego-cyclist (right) pairs.
We observe consistent shapes across datasets, where $60$--$80\%$ of interactions concentrate at 0, followed by a decaying tail.
Since $\mathrm{DRAC} = \Delta v^2 / 2d$ diverges as the longitudinal gap $d$ vanishes, we empirically clip it at $20$~m/s$^2$. The rightmost peak represents this saturation point, and holds less than $5\%$ of interactions per pair type.
The main difference is in \argoverse's pedestrian and cyclist distributions, which carry more mass across the mid-range, indicating ego interactions that close in faster.

\begin{figure}[!htbp]
    \centering
    \includegraphics[width=0.90\linewidth]{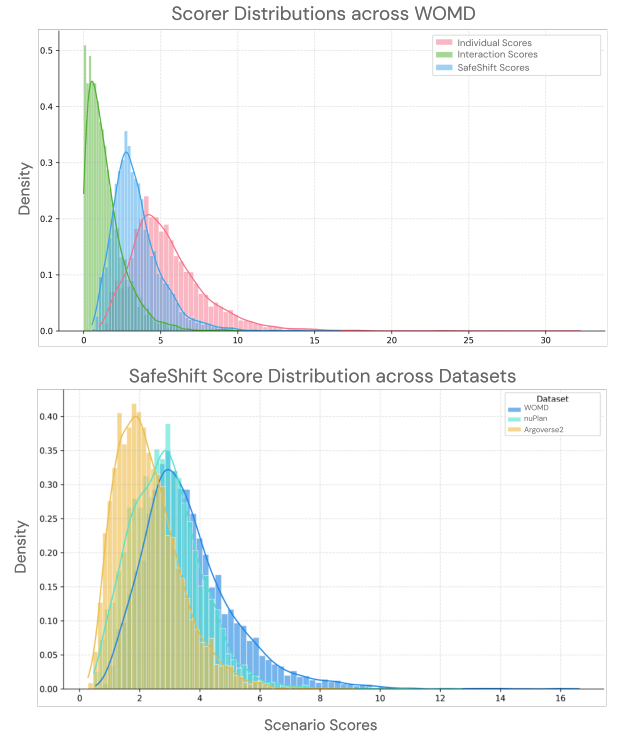}
    \caption{Score distribution analysis examples. \textbf{Top}: shows the score distribution across supported aggregators on \womd. \textbf{Bottom}: shows the \safeshift~distribution across \womd, \nuplan~and \argoverse.}
    \label{fig:score_analysis}
\end{figure}

\vspace{0.1cm}\subsubsection{Score analysis example}
The top subfigure in \Cref{fig:score_analysis} shows the score densities produced by the three currently supported aggregators (\Cref{ssec:scorer}) on \womd. All three densities are right-skewed, indicating that most scenes are nominal and that increasingly challenging scenarios sit in the tail. The interaction density is the most concentrated, peaking around 0 and largely exhausted after 5, reflecting that the majority of scenarios involve lower agent-to-agent interactivity. The individual density peaks near 4 and spreads past 15, indicating substantial variation in agents' kinematic profiles. Finally, the \safeshift~scorer represents a balance of the two, consistent with its formulation~\cite{stoler2024safeshift}.

The bottom subfigure compares the \safeshift~score distributions across datasets. While similarly shaped, they reveal differences in scenario composition and interestingness. For instance, \argoverse~concentrates at the lowest scores, \nuplan~sits in between, and \womd~shows the heaviest tail, extending past 10 where the others have decayed, consistent with its denser vehicle traffic (\Cref{tab:experimental_setup}).

\vspace{0.1cm}\subsubsection{Probe analysis example}
\begin{figure}[!htbp]
    \centering
    \includegraphics[width=0.92\linewidth]{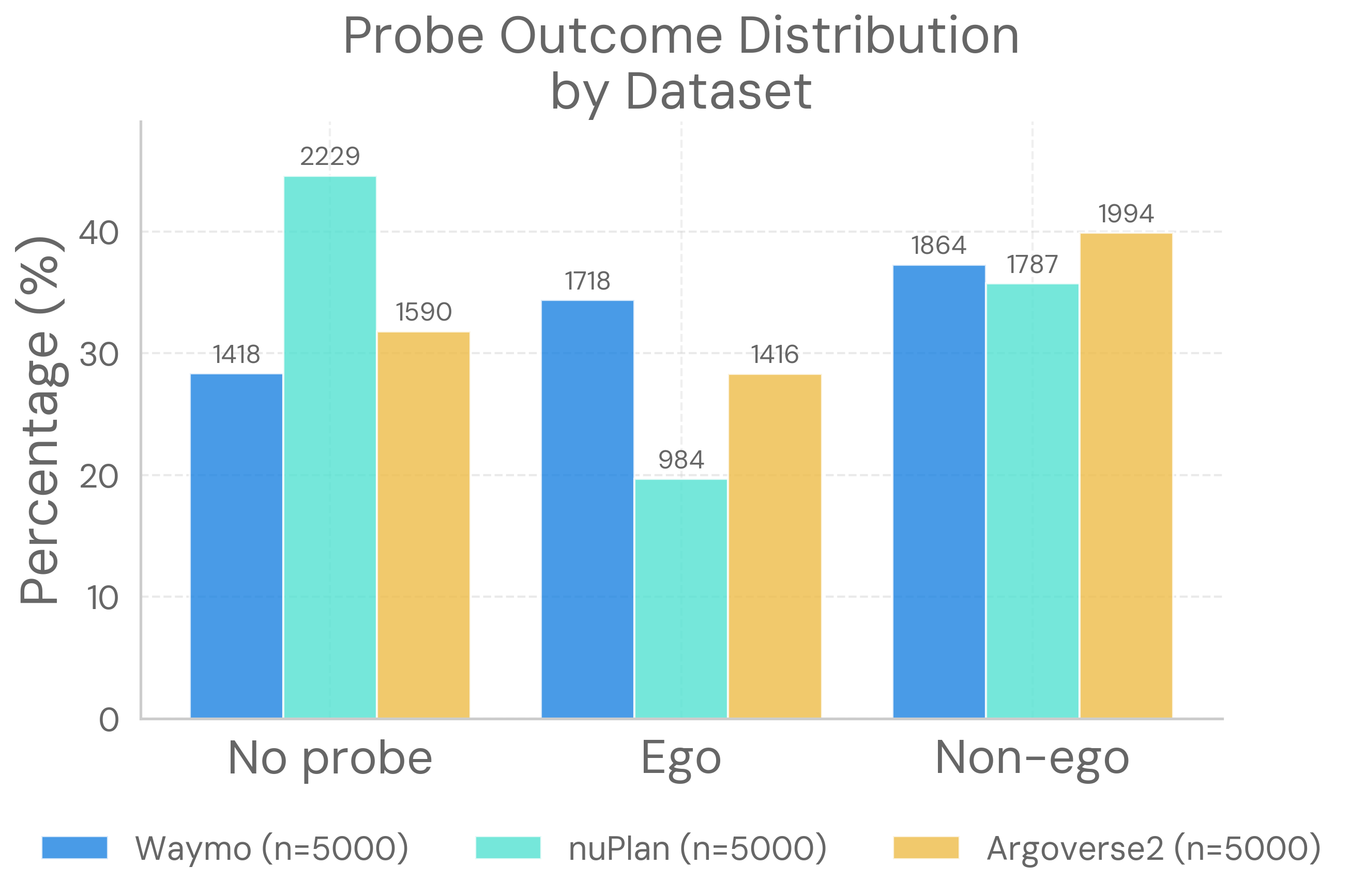}
    \caption{Probe outcome distribution across datasets. For each of the $5{,}000$ scenarios per dataset, the constant-velocity sweep either finds no critical probe or retains one that perturbs the \egoagent~or a non-ego agent; bars give the share of each outcome, annotated with raw counts.}
    \label{fig:probe_outcome_distribution}
\end{figure}

\Cref{fig:probe_outcome_distribution} reports how often the constant-velocity sweep finds no \textit{critical} probe in each dataset. It also reports how often the retained probe perturbs the \textit{ego} or a \textit{non-ego} agent. The non-ego rate is close across datasets ($36$--$40\%$). The ego rate is markedly different, at $34.4\%$ on \womd, $28.3\%$ on \argoverse, and $19.7\%$ on \nuplan. We hypothesize that vehicle traffic density (\Cref{tab:experimental_setup}) plays a role. A non-reactive ego is mostly consequential where a vehicle is close enough to be reached, \exempli in queues or at intersections. The gap may also indicate that the latent criticality exposed by simple non-reactive behavior is less prevalent in \nuplan. Richer probes (\Cref{ssec:future}), such as ones targeting VRUs, could help expose it.

\begin{figure*}[!htbp]
    \centering
    \includegraphics[width=0.98\textwidth]{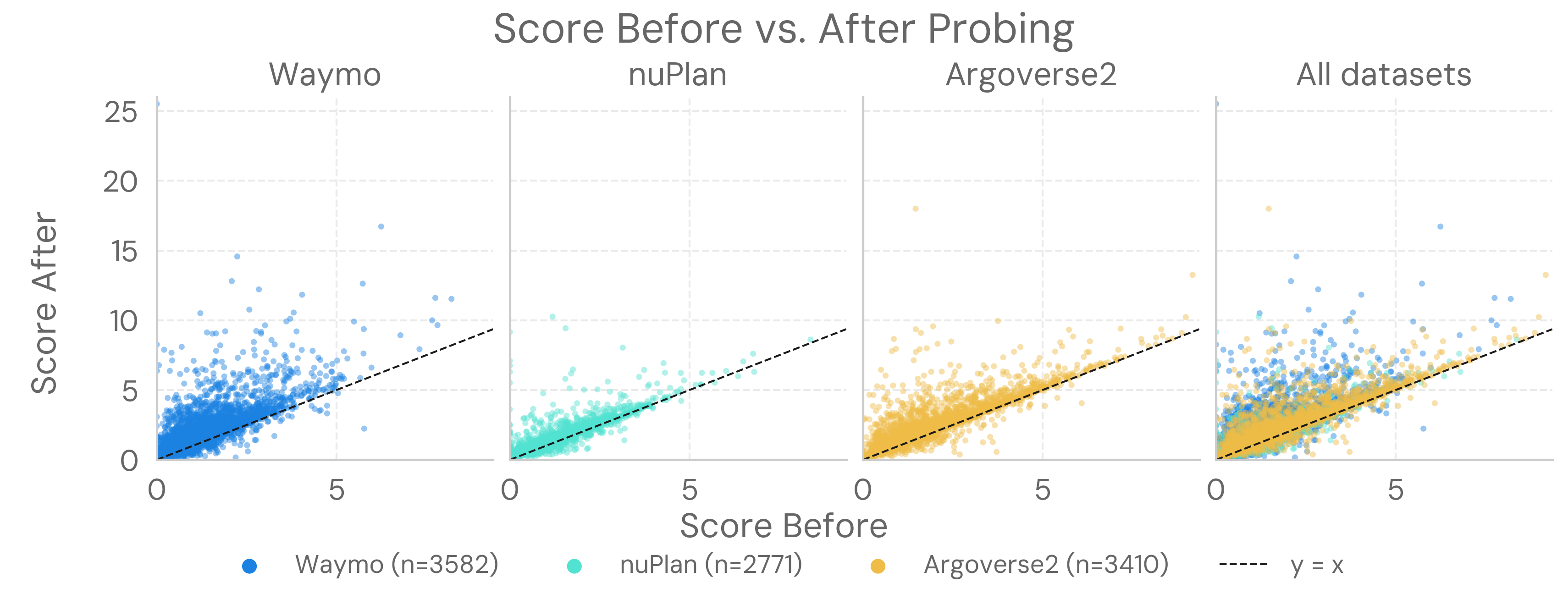}
    \caption{Per-scenario interaction score before vs.\ after probing, over the scenarios in which a probe was retained. \textbf{Left three}: per-dataset scatters for \womd, \nuplan, and \argoverse. \textbf{Rightmost}: all three overlaid. The dashed line marks $y=x$; vertical displacement from it is the probe-induced score delta.}
    \label{fig:scores_before_and_after}
\end{figure*}

\Cref{fig:scores_before_and_after} compares the per-scenario interaction score before and after probing, over the scenario pools with valid probes. The main cross-dataset difference is the \textit{scale} of the shift. \womd~shows the widest before-and-after gap, with many scenarios lifted well above the identity line. \argoverse~follows a similar but tighter pattern over a wider range of original scores. \nuplan's points stay close to the line. Even when a probe is found, it moves the score comparatively little, which reinforces the outcome distribution above.

\section{Downstream Applications}
\label{sec:applications}

The goal of this framework is to support downstream pipelines. We outline two illustrative directions enabled directly by the outputs of the framework as it stands.

\subsection{Agent Criticality Categorization}
The framework can be used to derive categorical labels, which can serve as auxiliary supervision and improve model interpretability. As an example, we designed a simple categorical profiler: it groups the raw descriptors by agent class for individual features and by pair type for interaction features, then partitions each (type, feature) distribution at $P$ configurable percentile thresholds into $P+1$ bands from low to high criticality (\Cref{fig:feature_analysis}). \Cref{fig:scenario_viz} shows example scenarios labeled by this strategy. The resulting per-agent labels can supervise behavior prediction or planning models, act as curriculum-learning anchors~\cite{bengio2009curriculum} that grade training data by difficulty, or flag agent relevance without manual annotation. A complementary direction is to feed the framework's scenario visualizations to large models, using their semantic reasoning to tag relevant agents, as in recent work on language-guided hard-case detection~\cite{yang2024hard}.

\subsection{Scenario Mining} 
\begin{figure}[!h]
    \centering
    \includegraphics[width=0.98\linewidth]{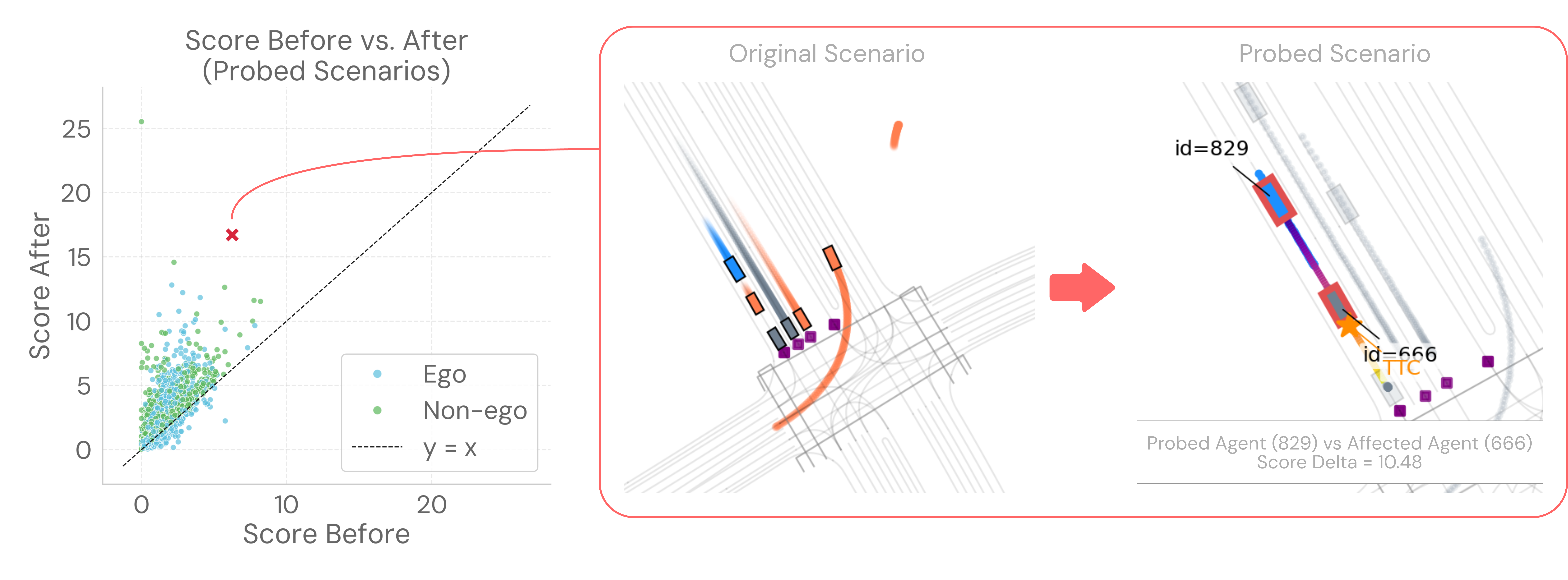}
    \caption{Mining scenarios by probe-induced criticality delta. \textbf{Left}: per-scenario scores before vs. after probing, color-coded by whether the probed agent is the ego or not. Points above the identity line are amplified by the probe. \textbf{Right}: the interaction marked with a red $\times$ in the scatter, shown before (left) and after (right) probing, where the constant-velocity probe raises the pair score by $10.5$.}
    \label{fig:scenario_mining}
\end{figure}

Because the framework produces per-scenario scores along multiple axes, the resulting distributions support mining at any point along the criticality spectrum. The tail of each density (\Cref{fig:score_analysis}) surfaces the most critical recorded scenarios, which can be held out for stress-testing prediction and planning models~\cite{stoler2024safeshift} or repurposed as \textit{templates} for adversarial generation~\cite{ding2023survey, huang2025cadre, stoler2025seal}. The counterfactual probing component (\Cref{ssec:prober}) extends this loop. By perturbing observed trajectories and re-scoring the resulting scenarios, the prober reshapes the original density (\Cref{fig:scores_before_and_after}), exposing latent criticality that recorded behavior alone does not reveal. Mining the perturbed densities (\Cref{fig:scenario_mining}), rather than the raw ones, could surface a more informative tail: scenarios that appear nominal as recorded but become safety-critical under small, plausible behavioral deviations. Investigating whether such scenarios correspond to cases where current models fail silently is a promising direction for future analysis.

\section{Discussion} \label{sec:discussion}

\subsection{Conclusion}
\label{ssec:conclusion}

We introduced \framework, an open-source, dataset-agnostic framework for automated characterization of driving scenarios. It generalizes the characterization and scoring methodologies of prior work~\cite{feng2024unitraj, stoler2024safeshift, glasmacher2022automated}, lifting scenario analysis out of a single-benchmark setting and packaging it as a reusable library that turns heterogeneous trajectory datasets into structured criticality signals available for visualization, mining, and categorization. We showcase the framework on three key AD datasets: \womd~\cite{ettinger2021large}, \argoverse~\cite{wilson2021argoverse}, and \nuplan~\cite{karnchanachari2024nuplan}.

\subsection{Limitations and Future Work}
\label{ssec:future}

Beyond the downstream uses explored in \Cref{sec:applications}, several direct extensions and improvements of this work are of interest. 
First, the current feature taxonomy is centered on agent kinematics and pairwise interactions; extending it with axes that capture purely environmental aspects of a scene, along with descriptors targeting sensor-level phenomena such as noisy or occluded observations, is a promising direction.

Second, the counterfactual-probing component was only preliminarily explored in this work. Future directions can incorporate a more extensive family of probes~\cite{ding2025surprise} and \textit{rule-based probes and validators}, which would test scenarios against explicit traffic rules and safety constraints.

Lastly, this characterization pipeline is, in principle, applicable beyond autonomous driving, but we have only demonstrated it on driving data. Extending the framework to adjacent domains such as aviation~\cite{navarro2024amelia, navarro2024ameliatf, patrikar2022trajair, navarro2022social}, or to the human-motion datasets exposed by unified interfaces such as \trajdata~\cite{ivanovic2023trajdata}, would test how far the abstraction carries. Together, these directions push the framework from a characterization tool toward a broader instrument for evaluating, mining, and generating safety-relevant trajectory data.


\addtolength{\textheight}{0cm}  

\section*{ACKNOWLEDGMENT} 
This work was performed during Ingrid Navarro's internship at Stack AV; we thank them for their mentorship and support throughout the process. 


\bibliographystyle{IEEEtran}
\bstctlcite{IEEEexample:BSTcontrol}
\bibliography{ref}

\end{document}